%% file: main_new.tex
\documentclass{bmvc2k}
\usepackage{times}
\usepackage{epsfig}
\usepackage{graphicx}
\usepackage{amsmath}
\usepackage{amssymb}
\usepackage{color}

\usepackage{bm}
\usepackage{hhline}
\usepackage{color}
\usepackage{booktabs}
\usepackage{url}
\usepackage{threeparttable}
\usepackage{multirow}
\newcommand{\qy}[1]{\textcolor{black}{#1}}
\newcommand{\unclear}[1]{\textcolor{black}{#1}}
\newcommand{\rg}[1]{\textcolor{black}{#1}}
\newcommand{\frg}[1]{\textcolor{black}{#1}}
\newcommand{\nickname}{XGrad-CAM}

\title{Axiom-based Grad-CAM: Towards Accurate Visualization and Explanation of CNNs}

\addauthor{Ruigang Fu}{furuigang08@nudt.edu.cn}{1}
\addauthor{Qingyong Hu}{qingyong.hu@cs.ox.ac.uk}{2}
\addauthor{Xiaohu Dong}{dongxiaohu16@nudt.edu.cn}{1}
\addauthor{Yulan Guo}{yulan.guo@nudt.edu.cn}{1}
\addauthor{Yinghui Gao}{yhgao@nudt.edu.cn}{1}
\addauthor{Biao Li}{libiao_cn@163.com}{1}

\addinstitution{
 College of Electronic Science and Technology\\
 National University of Defense Technology\\
 Changsha, China
}
\addinstitution{
 Department of Computer Science\\
 University of Oxford\\
 Oxfordshire, UK
}

\runninghead{Fu et al.}{Axiom-based Grad-CAM}

\begin{document}

\maketitle

\begin{abstract}
{To have a better understanding and usage of Convolution Neural Networks (CNNs), the visualization and interpretation of CNNs has attracted increasing attention in recent years. In particular, several Class Activation Mapping (CAM) methods have been proposed to discover the connection between CNN's decision and image regions. However, in spite of the reasonable visualization, most of these methods lack clear and sufficient theoretical support. \frg{In this paper, we introduce two axioms -- \emph{Sensitivity} and \emph{Conservation} -- to the visualization paradigm of the CAM methods. Meanwhile, a dedicated Axiom-based Grad-CAM (XGrad-CAM) is proposed to satisfy these axioms as much as possible.} Experiments demonstrate that XGrad-CAM is an enhanced version of Grad-CAM in terms of sensitivity and conservation. It is able to achieve better visualization performance than Grad-CAM, while also be class-discriminative and easy-to-implement compared with Grad-CAM++ and Ablation-CAM.} Code is available at \url{https://github.com/Fu0511/XGrad-CAM}.


\end{abstract}

\section{Introduction}
\label{sec:intro}
\qy{\frg{Due to the strong capability of feature learning, CNN-based approaches have achieved the state-of-the-art performance in numerous vision tasks such as image classification \cite{Krizhevsky2012ImageNet,Simonyan2014Very,He2016Identity}, object detection \cite{Ren2015Faster,Kaiming2017Mask} and semantic segmentation \cite{long2015fully}}. However, the interpretability of CNNs is often criticized by the community, as these networks usually look like complicated black boxes with massive unexplained parameters \cite{su2019one, dong2019efficient, li2019adversarial}. Therefore, it is highly desirable and necessary to find a way to understand and explain what exactly CNNs learned, especially for applications where interpretability is essential (e.g., medical diagnosis and autonomous driving). }


\rg{An important issue in CNN learning is to explain why classification CNNs predict what they predict \cite{Selvaraju2017Grad}. Since both semantic and spatial information can be preserved in feature maps of deep layers, Gradient-weighted Class Activation Mapping (Grad-CAM) \cite{Selvaraju2017Grad} was proposed to highlight important regions of an input image for CNN's prediction using deep feature maps.}
\qy{ Specifically, \frg{Grad-CAM} is defined as a linear combination of feature maps, where the weight of each feature map is determined by the average of its gradients. This definition is inspired by CAM \cite{Zhou2016Learning} and further improved by other works, such as Grad-CAM++ \cite{Aditya2017Grad} and Ablation-CAM \cite{ramaswamy2020ablation}. However, most of these CAM methods lack clear theoretical support, e.g., why does Grad-CAM \cite{Selvaraju2017Grad} use the average of gradients as the weight of each feature map? }

\begin{figure*}
\setlength{\belowcaptionskip}{-0.5cm} 
\begin{center}
\fbox{\parbox{8cm}{\hspace{5.5mm}\includegraphics[width=7cm]{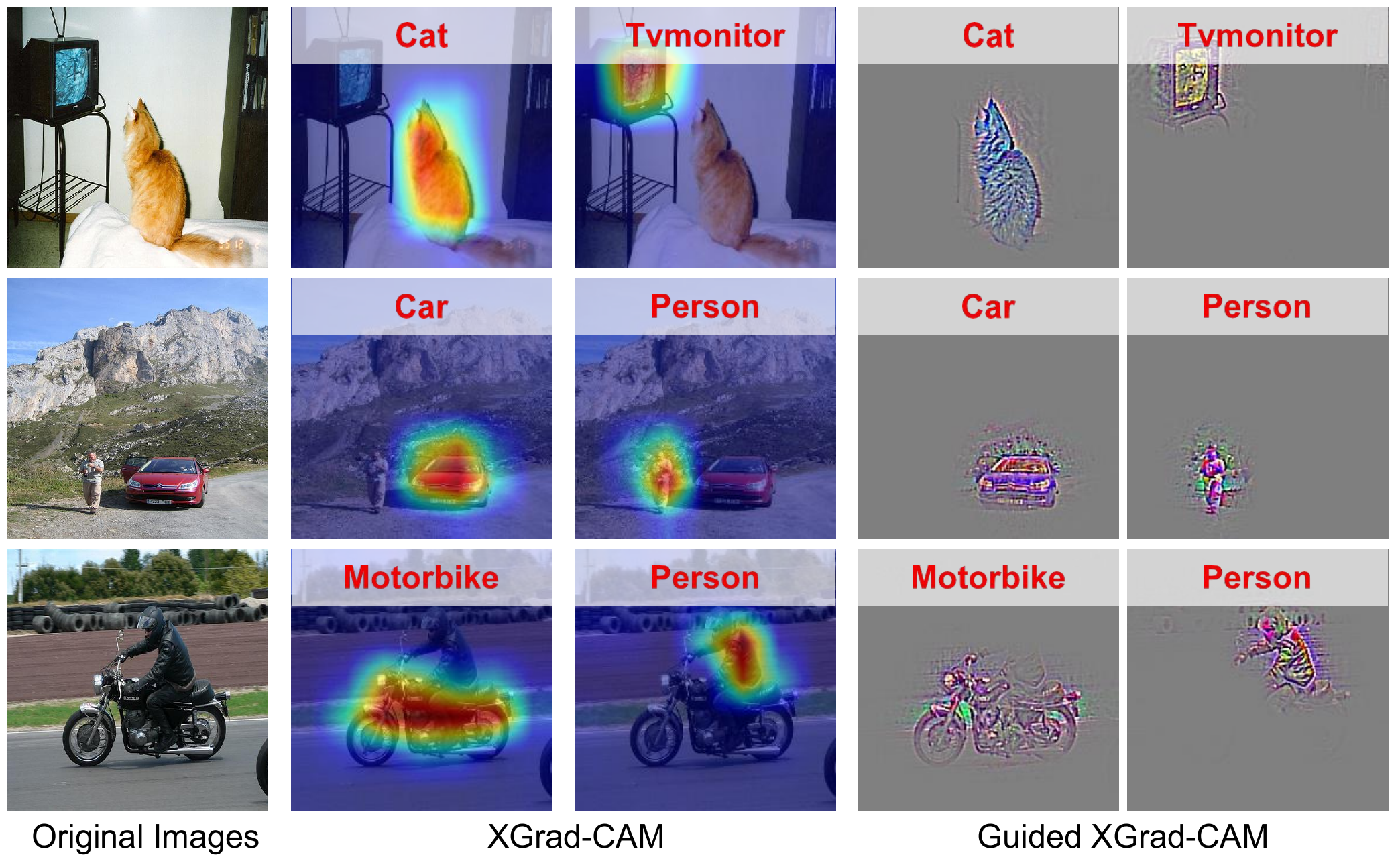}}}
\end{center}
   \caption{\qy{The visualization of our XGrad-CAM and Guided XGrad-CAM. It is clear that both of these two approaches are class-discriminative and able to highlight the object of interest. In addition, Guided XGrad-CAM provides more details than XGrad-CAM.}}
\label{fig:fig1}
\end{figure*}

In this paper, we propose a novel CAM method named XGrad-CAM (Axiom-based Grad-CAM) motivated by several formalized axioms. \frg{To achieve better visualization and explanation of CNN's decision, axioms are self-evident properties that visualization methods ought to satisfy \cite{montavon2018methods,sundararajan2017axiomatic, samek2019explainable}. Meeting these axioms makes a visualization method more reliable and theoretical. Therefore, two axiomatic properties are introduced in the derivation of XGrad-CAM: \emph{Sensitivity} \cite{sundararajan2017axiomatic} and \emph{Conservation} \cite{montavon2018methods}.} In particular, the proposed XGrad-CAM is still \frg{a linear combination of feature maps}, but able to meet the constraints of those two axioms. The weight of each feature map in XGrad-CAM is defined as a weighted average of its gradients by solving an optimization problem. As shown in Fig. \ref{fig:fig1}, our XGrad-CAM is a class-discriminative visualization method and able to highlight the regions belonging to the objects of interest. Further, by combining XGrad-CAM with the Guided Backprop \cite{Springenberg2014Striving}, we propose Guided XGrad-CAM which provides more details of the objects than XGrad-CAM.

\qy{To summarize, the main contributions of this work are as follows:}

\begin{itemize}
\item \qy{A dedicated \nickname{} with clear mathematical explanations is proposed to achieve better visualization of CNN's decision. It is able to be applied to arbitrary classification CNNs to highlight the objects of interest. }

\item \frg{By introducing two axioms as well as the corresponding axiom analysis, we can have a deeper understanding of why CAM methods work in visualizing the CNN's decision. Our XGrad-CAM can be seen as an enhanced version of Grad-CAM in both sensitivity and conservation. }

\item \qy{Extensive experiments have been conducted to give a comprehensive comparison between the proposed \nickname{} and several recent CAM methods (i.e., Grad-CAM \cite{Selvaraju2017Grad}, Grad-CAM++ \cite{Aditya2017Grad} and Ablation-CAM \cite{ramaswamy2020ablation}).  Taking both the class discriminability, efficiency and localization capability into consideration, our \nickname{} achieves better visualization performance.}



\end{itemize}

\section{Related Work}
\rg{A number of methods have started to visualize the internal representations learned by CNNs \cite{qin2018convolutional, montavon2018methods, guidotti2019survey} recently. These methods can be broadly categorized as: 1) visualization of filters and layer activations \cite{Krizhevsky2012ImageNet, Girshick2013Rich}, 2) visualization of hidden neurons \cite{simonyan2013deep,zeiler2014visualizing,mahendran2016visualizing,bau2017network}, 3) visualization of CNN's decision \cite{zeiler2014visualizing, simonyan2013deep, Zhou2016Learning, Selvaraju2017Grad}. In this section, we mainly introduce the visualization of CNN’s
decision and some related axioms.}

\subsection{Visualization of CNN's Decision}
\rg{These methods are developed to highlight the regions of an image, which are responsible for CNN's decision. They can be further categorized as: perturbation-based, propagation-based and \rg{activation-based} methods.}

\textbf{(1) Perturbation-based methods.} Zeiler et al. \cite{zeiler2014visualizing} occluded patches of an image using grey squares, and recorded the change of the class score. A heatmap can then be generated to show evidence for and against the classification. This method is further extended \cite{zintgraf2017visualizing, zhou2018efficient} using different types of perturbations such as removing, masking or altering. While perturbation-based methods are straightforward, they are inefficient.

\textbf{(2) Propagation-based methods.}
Propagation-based methods are rather fast, e.g., saliency maps proposed by Simonyan et al. \cite{simonyan2013deep} only require one forward propagation and one backward propagation through the network. Specifically, saliency maps use gradients to visualize relevant regions for a given class. However, with vanilla gradients, the generating saliency maps are usually noisy. Subsequent methods were developed to produce better visual heatmaps by modifying the back-propagation algorithm (e.g., Guided Backprop \cite{Springenberg2014Striving}, Layerwise Relevance Propagation \cite{bach2015pixel}, DeepTaylor \cite{montavon2017explaining}, Integrated Gradient\cite{sundararajan2017axiomatic}, etc.) or averaging the gradients for an input with noise added to it \cite{smilkov2017smoothgrad}.

\textbf{(3) \rg{Activation-based} methods.}
\qy{In contrast to propagation-based methods, \rg{activation-based} methods highlight objects by resorting to \rg{the activation of} feature maps. As an important branch of activation-based methods, \rg{CAM methods \cite{Selvaraju2017Grad, Aditya2017Grad, omeiza2019smooth, ramaswamy2020ablation} visualize CNN's decision using feature maps of deep layers.}
Zhou et al. \cite{Zhou2016Learning} proposed the original CAM method which visualizes a CNN by linearly combining feature maps at the penultimate layer. The weight of each feature map is determined by the last layer's fully-connected weights corresponding to a target class. However, CAM is restricted to GAP-CNNs. That is, the penultimate layer is constrainted to be a global average pooling (GAP) layer. Selvaraju et al. \cite{Selvaraju2017Grad} then proposed Grad-CAM to visualize an arbitrary CNN for classification by weighting the feature maps using gradients. Grad-CAM is inspired by CAM but hasn't explained its mechanism clearly (i.e., why using the average of gradients to weight each feature map). Aditya et al. \cite{Aditya2017Grad} proposed Grad-CAM++ by introducing higher-order derivatives in Grad-CAM. They assumed that the class score is a linear function of feature maps and got a closed-form solution of the weights for each feature map. Omeiza et al. \cite{omeiza2019smooth} further proposed Smooth Grad-CAM++. This method follows the framework of Grad-CAM++ but uses SmoothGrad  \cite{smilkov2017smoothgrad} to calculate gradients. More recently, Desai et al. \cite{ramaswamy2020ablation} proposed Ablation-CAM to remove the dependence on gradients but this method is quite time-consuming since it has to run forward propagation for hundreds of times per image.
}


\subsection{Axioms}
\label{sec::Axioms}
For a visualization method of CNN's decision, axioms are properties that are considered to be necessary for the method. Existing axioms include continuity \cite{montavon2018methods}, implementation invariance \cite{sundararajan2017axiomatic}, sensitivity \cite{sundararajan2017axiomatic} and conservation \cite{montavon2018methods}.

Given a model $m$, suppose that $d$ features constitute an input ${\bf x}$, $f({\bf x};m)$ represents a function of the model $m$ w.r.t the input ${\bf x}$. The resulting explanation is denoted by ${\bf R}({\bf x};m)\in \mathbb{R}^d$, where $R_i({\bf x};m)$ represents the importance of the $i$-th feature for the model output.

\textit{Continuity} is a property that if, for two nearly identical inputs, the model outputs are nearly identical, then the corresponding explanations should also be nearly identical, i.e., ${\bf R}({\bf x};m)\approx {\bf R}({\bf x}+{\bf \epsilon};m)$ with ${\bf \epsilon}$ a small perturbation.

\textit{Implementation invariance}. Two models $m_1$ and $m_2$ are functionally equivalent if they produce the same output for any identical input. Implementation invariance requires to produce identical explanations for functionally equivalent models provided with identical input, i.e., ${\bf R}({\bf x};m_1)={\bf R}({\bf x};m_2)$.

\textit{Sensitivity} is a property that each response of the explanation should be equal to the output change caused by removing the corresponding feature of the input, i.e., $R_i({\bf x};m)=f({\bf x};m)-f({\bf x}\backslash{x_i};m)$ where the notation ${\bf x}\backslash{x_i}$ indicates a modified input where the $i$-th feature in the original input has been replaced by a baseline value (usually zero).

\textit{Conservation} is a property that the sum of the explanation responses should match in magnitude of the model output, i.e., $f({\bf x};m)=\sum_{i=1}^d(R_i({\bf x};m))$.

In this paper, we mainly study the CAM methods using the axioms of sensitivity and conservation to visualize CNN's decision. Generally, gradient-based CAM methods violate the axiom of continuity because of the problem of shattered gradients \cite{samek2019explainable}. Besides, they also break the axiom of implementation invariance since they are layer sensitive \cite{ramaswamy2020ablation}.

\section{Approach}

\subsection{Notation and Motivation}
\qy{Given an $L$-layer CNN and an input image $\bf I$, let $l$ represent the index of the target layer for visualization, ${\bf F}^l$ denote the response of the target layer, $S_c({\bf F}^l)$ represent the class score (the input to the softmax layer) of a class of interest $c$. Suppose that the $l$-th layer contains $K$ feature maps, where the response of the $k$-th feature map is denoted as ${\bf F}^{lk}$. $F^{lk}(x,y)$ represents the response at position $(x,y)$ in ${\bf F}^{lk}$.
}


\qy{To visualize the class $c$ in the input image, a general form of the existing CAM methods \cite{Selvaraju2017Grad,Aditya2017Grad,ramaswamy2020ablation} can be written as a linear combination of the feature maps in the target layer:}

\begin{small}
\begin{equation}\label{eq1}
M_c(x,y)=\sum_{k=1}^K\left(w^k_c{F^{lk}(x,y)}\right),
\end{equation}
\end{small}

\noindent \qy{ where $w^k_c$ is the weight of the corresponding feature map ${\bf F}^{lk}$, different definitions lead to different CAM methods. \frg{Then, to further identify the image regions responsible for the particular class $c$, two postprocessing are needed: the resulting map $M_c$ needs to be ReLU rectified to filter negative units \cite{Selvaraju2017Grad} and upsampled to the same size of the input image.}}

For the CAM methods, the key problem is how to precisely determine the importance of each feature map to the prediction of the class of interest. \qy{In this paper, we argue that it would be better if these CAM methods can satisfy two basic axioms, i.e.,  sensitivity and conservation.}




{\bf Sensitivity:}
\qy{A general CAM method of Eq. \eqref{eq1} satisfies the axiom of sensitivity if it holds the following property for all the feature maps, that is:}

\begin{small}
\begin{equation}\label{eq3}
S_c({\bf F}^{l})-S_c({\bf F}^{l}\backslash{\bf F}^{lk})=\sum_{x,y}\left(w^k_c{F^{lk}(x,y)}\right),
\end{equation}
\end{small}

\noindent \frg{where $S_c({\bf F}^{l}\backslash{\bf F}^{lk})$ is the score of class $c$ when the $k$-th feature map in the target layer has been replaced by zero.} This means that the importance of each feature map should be equivalent to the score change caused by its removing.

{\bf Conservation:}
\qy{To meet the axiom of conservation, a general CAM method of Eq. \eqref{eq1} should hold:}

\begin{small}
\begin{equation}\label{eq2}
S_c({\bf F}^l)=\sum_{x,y}\left(\sum_{k=1}^K\left(w^k_c{F^{lk}(x,y)}\right)\right).
\end{equation}
\end{small}

\qy{This means that the responses of the CAM map should be a redistribution of the class score. }

Intuitively, if a large drop of class score appears when we removed a specific feature map, this feature map would be expected as high importance. Sensitivity is exactly an axiom based on this intuition. Besides, conservation is introduced here to ensure that the class score can be mainly dominated by the feature maps rather than other unexplained factors. Therefore, introducing sensitivity and conservation is likely to make the CAM methods achieve more reasonable visualization.

\qy{To meet the above two axioms as much as possible, we formulate this as a minimization problem of $\phi(w^k_c)$ as below: }


%

\begin{footnotesize}
\begin{equation}\label{eq4}
\begin{split}
\phi(w^k_c)=
\underbrace{\sum_{k=1}^K\left|S_c({\bf F}^{l})-S_c({\bf F}^{l}\backslash{\bf F}^{lk})-\sum_{x,y}\left(w^k_c{F^{lk}(x,y)}\right)\right|}_{\textrm{Sensitivity}}+
\underbrace{\mathop{\left|S_c({\bf F}^{l})-\sum_{x,y}\left(\sum_{k=1}^K\left(w^k_c{F^{lk}(x,y)}\right)\right)\right|}}_{\textrm{Conservation}}.
\end{split}
\end{equation}
\end{footnotesize}


\subsection{Our Method}

\qy{Given an arbitrary target layer in a ReLU-CNN which only has ReLU activation as its nonlinearities, we can prove that for any class of interest, the class score is equivalent to the sum of the element-wise products between feature maps and gradient maps of the target layer, followed with a bias:}


\begin{footnotesize}
\begin{equation}\label{eq5}
\begin{split}
S_c({\bf F}^{l})=\sum_{k=1}^K\sum_{x,y}\left(\frac{\partial{S_c({\bf F}^l)}}{\partial{F^{lk}(x,y)}}{F^{lk}(x,y)}\right)+\epsilon({\bf F}^l),
\end{split}
\end{equation}
\end{footnotesize}

\noindent \qy{where $\epsilon({\bf F}^l)=\sum_{t=l+1}^L\sum_{j}\frac{\partial{S_c({\bf F}^l)}}{\partial{u_j^{t}}}b_j^{t}$, $u_j^{t}$ denotes a unit in the layer $t$ ($t$>$l$) and $b_j^{t}$ is the bias term corresponding to the unit $u_j^{t}$.} The detailed proof can be found in Appendix.




\begin{figure}
\setlength{\belowcaptionskip}{-0.5cm}
\begin{tabular}{cc}\hspace{1.4mm}
\bmvaHangBox{\fbox{\parbox{5.5cm}{\includegraphics[width=5.5cm]{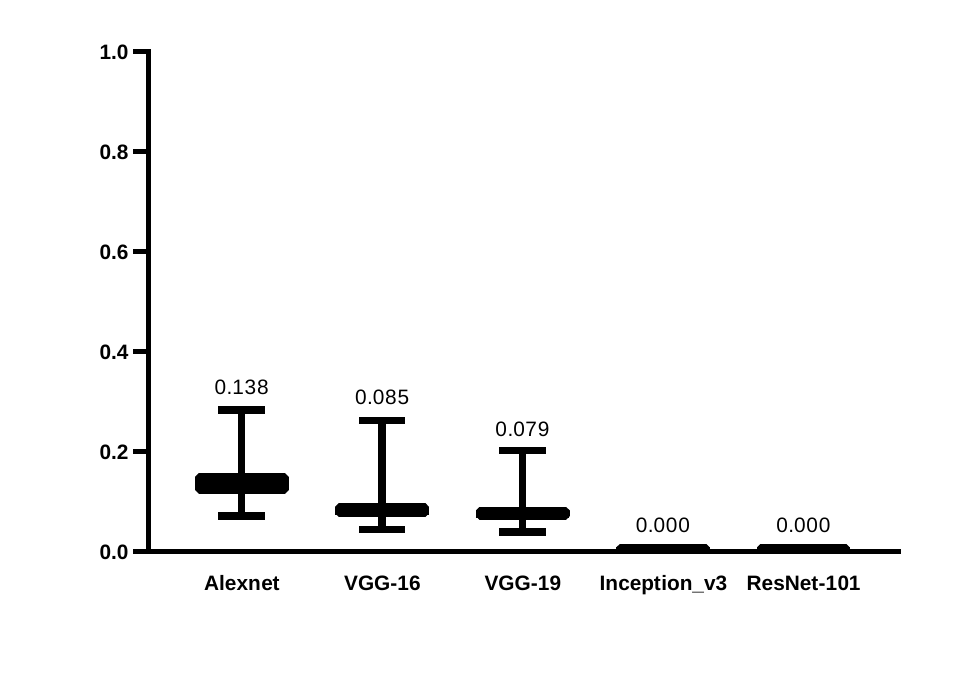}}}}&
\bmvaHangBox{\fbox{\parbox{5.5cm}{\includegraphics[width=5.5cm]{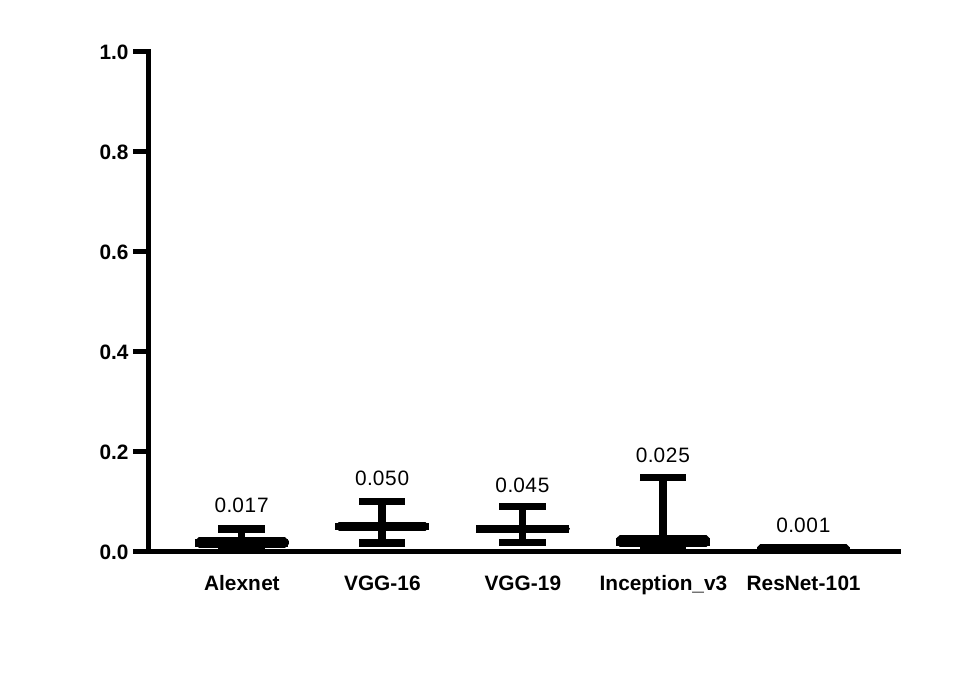}}}}\\
(a)&(b)
\end{tabular}
\caption{(a) Normalized $\zeta({\bf F}^{l};k)$ is small in the last spatial layers of different CNN models, including AlexNet \cite{Krizhevsky2012ImageNet}, VGG-16 \cite{Simonyan2014Very}, VGG-19 \cite{Simonyan2014Very}, Inception\_V3 \cite{Szegedy2016Rethinking} and ResNet-101 \cite{He2016Identity}; (b) Normalized $\epsilon({\bf F}^l)$ is also small in the last spatial layers of different CNN models. The mean values are provided above the box-plots.}
\label{fig:fig2}
\end{figure}

\qy{By further substituting the $S_c({\bf F}^{l})$ in Eq. \eqref{eq4} with the value in Eq. \eqref{eq5}, we can have:}

\begin{small}
\begin{equation}\label{eq6}
\begin{split}
\phi(w^k_c)=\sum_{k=1}^K\left|\sum_{x,y}\left(\frac{\partial{S_c({\bf F}^{l})}}{\partial{F^{lk}(x,y)}}F^{lk}(x,y)-w^k_cF^{lk}(x,y)\right)+\zeta({\bf F}^{l};k)\right|+\\
\left|\sum_{k=1}^K\sum_{x,y}\left(\frac{\partial{S_c({\bf F}^{l})}}{\partial{F^{lk}(x,y)}}{F^{lk}(x,y)}-w^k_c{F^{lk}(x,y)}\right)+\epsilon({\bf F}^l)\right|.
\end{split}
\end{equation}
\end{small}

\noindent where $\zeta({\bf F}^{l};k)=\sum_{k^{'}=1,k^{'} \neq k}^K\sum_{x,y}\left(\frac{\partial{S_c({\bf F}^{l})}}{\partial{F^{lk{'}}(x,y)}}F^{lk{'}}(x,y)-\frac{\partial{S_c({\bf F}^{l}\backslash{\bf F}^{lk})}}{\partial{F^{lk{'}}(x,y)}}F^{lk{'}}(x,y)\right)+\epsilon({\bf F}^l)-\epsilon({\bf F}^{l}\backslash{\bf F}^{lk})$. For the terms $\zeta({\bf F}^{l};k)$ and $\epsilon({\bf F}^l)$, they are difficult to optimize by the variable $w_c^k$ since there are no direct relationship between these terms and the $k$-th feature map of the target layer. As a workaround, we calculated their normalized versions (i.e.,
$\frac{\sum_k\left|\zeta({\bf F}^{l};k)\right|}{\sum_k\left|S_c({\bf F}^{l})-S_c({\bf F}^{l}\backslash{\bf F}^{lk})\right|}$ and $\left|\frac{\epsilon({\bf F}^l)}{S_c({\bf F}^{l})}\right|$)
of 1000 input images in the last spatial layers of several classical CNN models, with the class of interest $c$ set as the top-1 predicted class. We empirically found that the average of these terms are rather small for all the models as shown in Fig. ~\ref{fig:fig2}. In contrast, these terms are rather large in shadow layers, a visualization of $\epsilon({\bf F}^l)$ in different layers of VGG16 model is provided in Appendix. Therefore, when the target layer is deep, to minimize Eq. \eqref{eq6}, we can calculate an approximate optimal solution $\alpha_c^k$ by making the terms in $\left | \cdot  \right |$ equal to zero without considering $\zeta({\bf F}^{l};k)$\footnote{To be precise, $\frac{\sum_k\left|\zeta({\bf F}^{l};k)\right|}{\sum_k\left|S_c({\bf F}^{l})-S_c({\bf F}^{l}\backslash{\bf F}^{lk})\right|}$ is small can not indicate that $\frac{\left|\zeta({\bf F}^{l};k)\right|}{\left|S_c({\bf F}^{l})-S_c({\bf F}^{l}\backslash{\bf F}^{lk})\right|}$ is small for all the feature maps. Exceptions exist indeed but they usually happen in the unimportant feature maps whose removing only lead to a tiny score change. For those feature maps, $\zeta({\bf F}^{l};k)$ is still rather small and can be ignored because it has little influence on the final visualization. A visual example is provided in Appendix.} and $\epsilon({\bf F}^l)$:

\begin{small}
\begin{equation}\label{eq7}
\alpha^k_c=\sum_{x,y}\left(\frac{F^{lk}(x,y)}{\sum_{x,y}{F^{lk}(x,y)}}\frac{\partial{S_c({\bf F}^{l})}}{\partial{F^{lk}(x,y)}}\right).
\end{equation}
\end{small}


\qy{In this case, we define our XGrad-CAM by substituting $w^k_c$ in Eq. \eqref{eq1} with $\alpha^k_c$ :}


\begin{small}
\begin{equation}\label{eq9}
M_c^{\textrm{XGrad-CAM}}(x,y)=\sum_{k=1}^K\left(\alpha^k_c{F^{lk}(x,y)}\right).
\end{equation}
\end{small}

\begin{figure*}
\setlength{\belowcaptionskip}{-0.5cm} 
\begin{center}
\fbox{\parbox{9.7cm}{\hspace{0mm}\includegraphics[width=9.7cm]{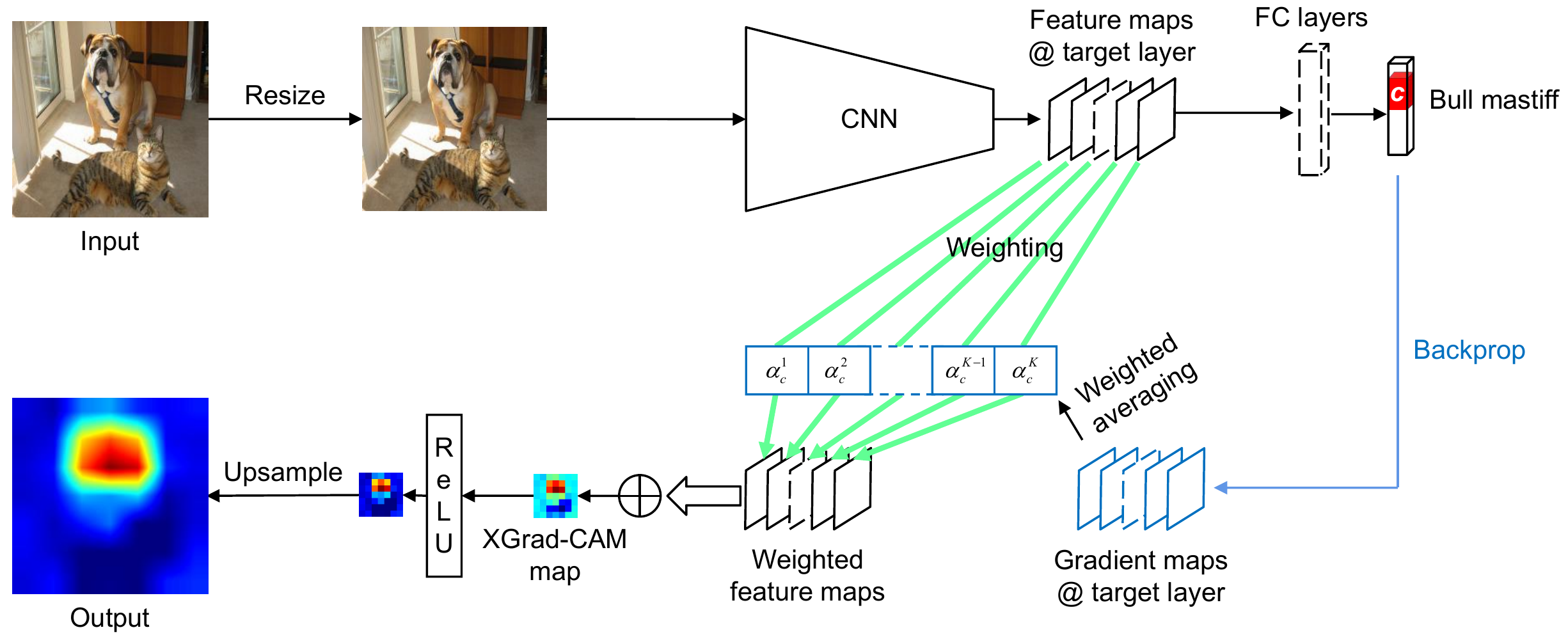}}}
\end{center}
\caption{An overview of the XGrad-CAM scheme.} 
\label{fig:fig10}
\end{figure*}

\noindent \frg{Then, by rectifying the resulting map using ReLU function and upsampling the map to the input size, we can identify the image regions responsible for the class $c$ as shown in Fig. ~\ref{fig:fig10}.}

\qy{It can be proved that XGrad-CAM is a generalization of CAM \cite{Zhou2016Learning} since it is identical to CAM for the GAP-CNNs but can be applied to arbitrary CNN models (see Appendix for the proof). Besides, we also propose Guided XGrad-CAM by multiplying the up-sampled XGrad-CAM by the Guided Backprop \cite{Springenberg2014Striving} element-wisely. }

\section{Experiments and Results}

\qy{In this section, we mainly evaluate the performance of different CAM methods, including Grad-CAM \cite{Selvaraju2017Grad}, Grad-CAM++ \cite{Aditya2017Grad}, Ablation-CAM \cite{ramaswamy2020ablation} and our XGrad-CAM. We argue that accurate localization of the objects of interest in an input image is necessary for an ideal visualization approach. Therefore, we evaluate the visualization quality from ``class-discriminability'' (see Sec. \ref{sec:class}) and ``localization capability'' (see Sec. \ref{sec:perturbation}). In addition, we also analyze the rationality of existing methods from the perspective of axiom in Sec. \ref{Axiom Analysis}.}

\subsection{Experimental Setup}
\qy{All of the existing methods are based on Eq. \eqref{eq1} but with different weights $w_c^k$, that is:}

\begin{itemize}
\item \qy{Grad-CAM. The weight of each feature map is defined as $\frac{1}{Z}\sum_{x,y}\frac{\partial{S_c({\bf F}^{l})}}{\partial{F^{lk}(x,y)}}$ where $Z$ is the number of units in the $k$-th feature map.}

\item \qy{Grad-CAM++. The weight of each feature map is defined as $\sum_{x,y}a^k_c(x,y)\textrm{ReLU}(\frac{\partial{S_c({\bf F}^{l})}}{\partial{F^{lk}(x,y)}})$ where $a^k_c(x,y)$ is a closed form weight based on an assumption that the class score is a linear function of feature maps.} 

\item \rg{Ablation-CAM. The weight of each feature map is defined as $\frac{S_c({\bf F}^{l})-S_c({\bf F}^{l}\backslash{\bf F}^{lk})}{\sum_{x,y}F^{lk}(x,y)}$, it removes the dependence of gradients. Note that the original weight of each feature map in Ablation-CAM \cite{ramaswamy2020ablation} is defined as $\frac{S_c({\bf F}^{l})-S_c({\bf F}^{l}\backslash{\bf F}^{lk})}{||{\bf F}^{lk}||}$. Since the selected target layer of CAM methods is usually ReLU rectified, the responses of the feature maps are always positive, we set $||{\bf F}^{lk}||$ as $\sum_{x,y}F^{lk}(x,y)$ in this paper\footnote{ \noindent $||{\bf F}^{lk}||$ is roughly set to $S_c({\bf F}^{l})$ in the original paper. }. It is easy to verify that Ablation-CAM here totally satisfies the axiom of sensitivity.}

    %
\end{itemize}

\frg{These CAM methods are performed on the last spatial layer of VGG-16 model \cite{Simonyan2014Very} pre-trained on the ImageNet. For GAP-CNNs (e.g., ResNet-101 \cite{He2016Identity} and Inception\_V3 \cite{Szegedy2016Rethinking}), it can be proved that Grad-CAM, Ablation-CAM and XGrad-CAM achieve the same performance on the last spatial layers of the models (refer to Appendix for the detailed proof). All experiments are implemented in Pytorch \cite{paszke2017automatic} and conducted on an NVIDIA Titan Xp GPU.}


\begin{figure}
\begin{tabular}{cc}
\bmvaHangBox{\fbox{\parbox{4.4cm}{\includegraphics[width=4.4cm]{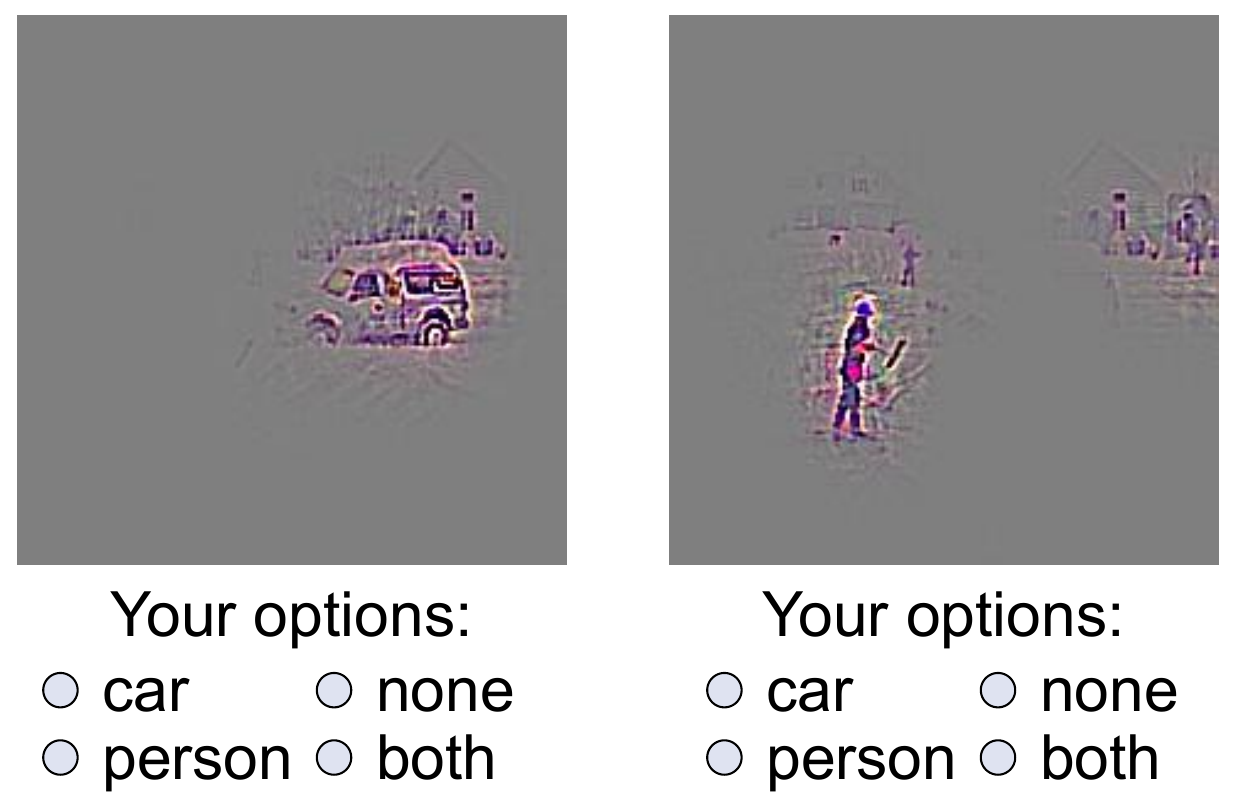}}}}&
\bmvaHangBox{\fbox{\parbox{7.4cm}{\includegraphics[width=7.4cm]{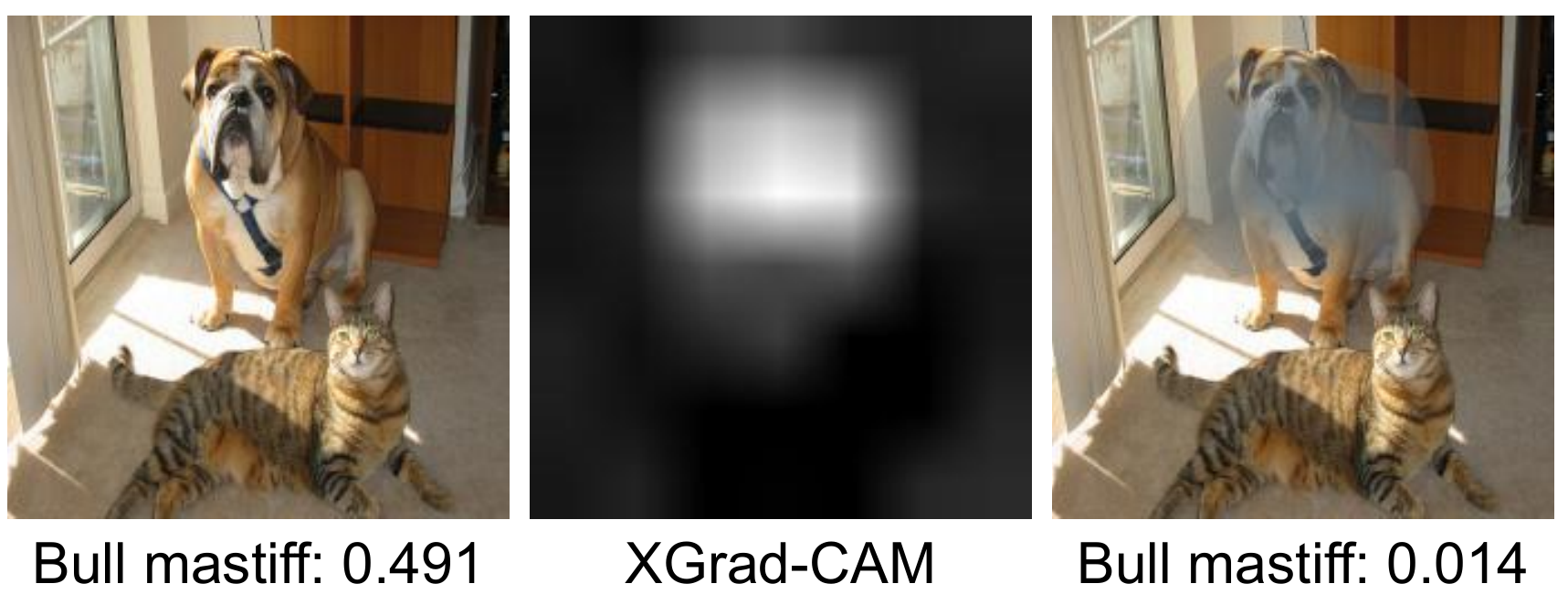}}}}\\
(a)&(b)
\end{tabular}
\caption{(a) A game of ``What do you see'' to evaluate the class-discriminability of each CAM method. Subject needs to answer what is being depicted in the visualization; (b) An example of XGrad-CAM visualization and its corresponding perturbed image. }
\label{fig:fig5}
\end{figure}

\begin{table}[tb]\small
\setlength{\abovecaptionskip}{-3pt}
\setlength{\belowcaptionskip}{-0.5cm} 
\centering
\begin{threeparttable}
{\centering
\begin{tabular}{|c|c|c|c|}
\hline
 {Method}&{Class\_discrimination}&{Confidence\_drop}&{Efficiency (s)}\\
\hline
\hline
 {{Grad-CAM} \cite{Selvaraju2017Grad}}&{\bf $\approx$0.709}&{0.469}&{\bf 0.021}\\
 {{Grad-CAM++}\cite{Aditya2017Grad}}&{$\approx$0.308}&{\bf 0.494}&{0.022}\\
 {{Ablation-CAM} \cite{ramaswamy2020ablation}}&{$\approx$0.700}&{0.484}&{0.735}\\
 {{XGrad-CAM}}&{$\approx$0.702}&{0.491}&{\bf 0.021}\\
\hline
\end{tabular}}
\end{threeparttable}
\caption{Results of class discrimination analysis and perturbation analysis. It is shown that Grad-CAM++ \cite{Aditya2017Grad} is not class-discriminative, Grad-CAM \cite{Selvaraju2017Grad} is not good enough in localizing the object of interest, Ablation-CAM \cite{ramaswamy2020ablation} is time-consuming.}\label{tab:tab2}
\end{table}

\subsection{Class Discrimination Analysis}
\label{sec:class}
\qy{A good visualization of CNN's decision should be class-discriminative. Specifically, the visualization method should only highlight the object belonging to the class of interest in an image when there are objects labeled with several different classes.}


\qy{To evaluate the ability of class discrimination, we followed the subjective evaluation method used in \cite{Aditya2017Grad}. Specifically, we first finetuned the pre-trained VGG-16 model on the Pascal VOC 2007 training set. The images in VOC set usually contain multiple objects belonging to different classes. We then selected 100 images from VOC 2007 validation set that contain exactly two classes. For each image and each CAM method, we used the guided version of the CAM method to generate a pair of class-specific visualizations as shown in Fig. \ref{fig:fig5} (a). These visualizations were then shown to 5 individuals who were asked to answer a choice question: what class is highlighted by the visualizations. Note that the options also include ``none'' and ``both'' which are both incorrect answers.}

\qy{Quantitative results are shown in Table \ref{tab:tab2}. The class-discriminative evaluation is subjective, but it is clear that the performance of Grad-CAM, Ablation-CAM and XGrad-CAM are very similar. On the other hand, Grad-CAM++ performs much worse than the other three methods. For more visualization results and the reason why we evaluated the class-discriminability of different CAM methods using their guided versions rather than themselves, please refer to the Appendix. }

\begin{figure*}[t]
\setlength{\belowcaptionskip}{-0.5cm} 
\begin{center}
\fbox{\parbox{12.5cm}{\includegraphics[width=12.5cm]{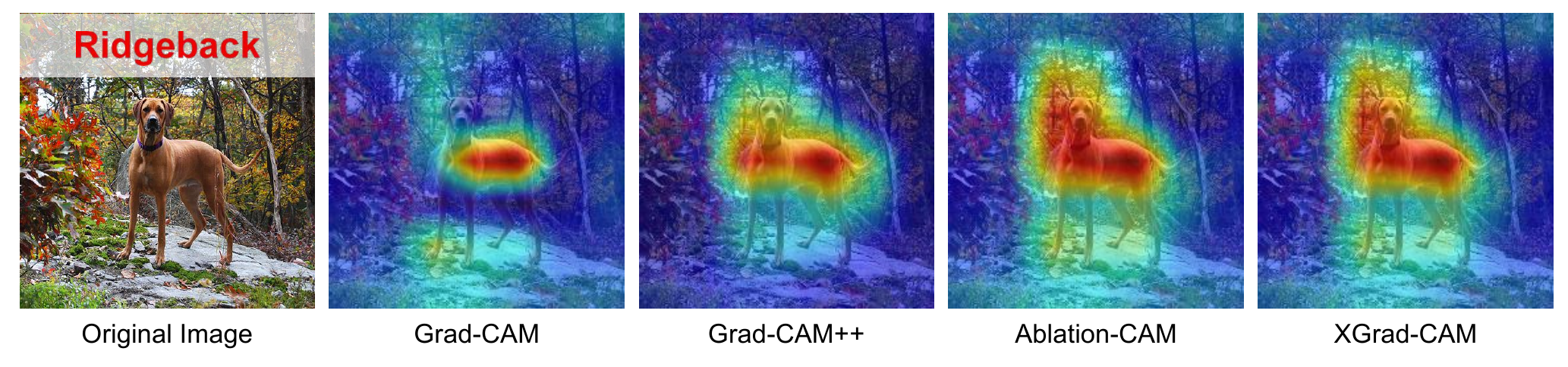}}}
\end{center}
   \caption{Example explanation maps generated by Grad-CAM \cite{Selvaraju2017Grad}, Grad-CAM++ \cite{Aditya2017Grad}, Ablation-CAM \cite{ramaswamy2020ablation} and our XGrad-CAM.}
\label{fig:fig3}
\end{figure*}

\subsection{Perturbation Analysis}
\label{sec:perturbation}

\qy{The localization capability of CAM methods is usually evaluated by perturbation analysis \cite{Selvaraju2017Grad,Aditya2017Grad,ramaswamy2020ablation}. The underlying assumption is that the perturbation of relevant regions in an input image should lead to a decrease in class confidence.}

\qy{We followed the evaluation scheme used in \cite{Aditya2017Grad,ramaswamy2020ablation} for perturbation analysis. The experiments were conducted on the ILSVRC-12 validation set \cite{russakovsky2015imagenet}. Take XGrad-CAM for example, each image ${\bf I}_i$ in the dataset was first fed to the VGG-16 model to predict its top-1 class. Then, XGrad-CAM method was used to generate a corresponding heatmap ${\bf H}_i$ for the predicted class. Inspired by the meaningful perturbation illustrated in \cite{fong2017interpretable}, we perturbed the original image by masking the regions highlighted by the XGrad-CAM method:
}


\begin{small}
\begin{equation}\label{eq12}
\widetilde{\bf I}_i={\bf I}_i \circ(1-{\bf M}_i)+\mu{\bf M}_i,
\end{equation}
\end{small}

\noindent \qy{where ${\bf M}_i$ is a mask based on the original heatmap ${\bf H}_i$. Specifically, in the mask, only the pixels corresponding to the top 20\% value of the heatmap are set equal to the heatmap, while the rest are set to 0. ``$\circ$'' represents the element-wise multiplication and $\mu$ is the mean value used in the input normalization. A perturbed example is shown in Fig. \ref{fig:fig5} (b). It is shown that with the perturbation, the confidence of the target class ``Bull mastiff'' decreases sharply. }


\qy{Then, we computed the difference of the class confidence (the output of the softmax layer) between the original image and the perturbed image:}
\begin{small}
\begin{equation}\label{eq13}
Confidence\_drop=\frac{1}{N}\sum_{i=1}^{N}{\frac{P_c({\bf I}_i)-P_c(\widetilde{\bf I}_i)}{P_c({\bf I}_i)}},
\end{equation}
\end{small}

\noindent \frg{where $P_c({\bf I}_i)$ and $P_c(\widetilde{\bf I}_i)$ are the class confidences of the original image ${\bf I}_i$ and perturbed image $\widetilde{\bf I}_i$, respectively}, $N$ is the total number of images in dataset. If the heatmap has highlighted the regions that are most important for class $c$, the confidence drop is expected to be larger.

\qy{The results are shown in Table \ref{tab:tab2}, we can see that XGrad-CAM achieves better performance than Grad-CAM (0.491 v.s. 0.469). Ablation-CAM performs similar to XGrad-CAM, but it is much more time-consuming than XGrad-CAM (about 40 times), it has to run hundreds of forward propagation per image. While Grad-CAM++ achieves the best performance, its class-discriminability is lost (refer to Section \ref{sec:class}). Note that, the class-discriminability cannot be reflected by the confidence drop on the ILSVRC-12 validation set because images in this dataset usually contain a single object class. Visual example results generated by the different CAM methods are shown in Fig. \ref{fig:fig3}, it can be observed that the result achieved by XGrad-CAM covers a more complete area of the object than Grad-CAM.}

\frg{To summarize,} Grad-CAM++ \cite{Aditya2017Grad} is not class-discriminative, Ablation-CAM \cite{ramaswamy2020ablation} is time-consuming, Grad-CAM \cite{Selvaraju2017Grad} is not good enough in localizing the object of interest. Therefore, considering the property of class discrimination, efficiency and the localization performance \frg{comprehensively}, our XGrad-CAM is a promising visualization scheme in practice.




\begin{table}[tb]\small
\setlength{\abovecaptionskip}{-3pt}
\setlength{\belowcaptionskip}{-0.5cm} 
\centering
\begin{threeparttable}
{\centering
\begin{tabular}{|c|c|c|}
\hline
 {Method}&{Sensitivity}&{Conservation}\\
 \hline\hline
 {{Grad-CAM} \cite{Selvaraju2017Grad}}&{0.313}&{0.303}\\
 {{Grad-CAM++}\cite{Aditya2017Grad}}&{$\gg$1}&{$\gg$1}\\
 {{Ablation-CAM} \cite{ramaswamy2020ablation}}&{\bf 0}&{0.145}\\
 {{XGrad-CAM}}&{0.085}&{\bf 0.051}\\
\hline
\end{tabular}}
\end{threeparttable}
\caption{Results of axiom analysis on the ILSVRC-12 validation set when applying CAM methods on the last spatial layers of VGG-16 model.}\label{tab:tab1}
\end{table}
\subsection{Axiom Analysis}
\label{Axiom Analysis}
\qy{To further study whether the existing CAM methods satisfy the axioms of sensitivity and conservation, we conduct axiom analysis on the ILSVRC-12 validation set.} Specifically, the sensitivity of a general CAM method of Eq. \eqref{eq1} can be measured by: 

\begin{small}
\begin{equation}\label{eq11}
\frac{1}{N}\sum_{i=1}^{N}\frac{\sum_{k=1}^{K}{|S_c({\bf F}^{l}_i)-S_c({\bf F}^{l}_i\backslash{\bf F}^{lk}_i)-\sum_{x,y}\left(w^k_c{F^{lk}_i(x,y)}\right)|}}{\sum_{k=1}^{K}{|S_c({\bf F}^{l}_i)-S_c({\bf F}^{l}_i\backslash{\bf F}^{lk}_i)|}}.
\end{equation}
\end{small}

\noindent \qy{where ${\bf F}^{l}_i$ is the response of the target layer of the $i$-th image in the dataset, $c$ is the top-1 class predicted by the VGG-16 model. Analogously, the conservation can be measured by:}

\begin{small}
\begin{equation}\label{eq10}
\frac{1}{N}\sum_{i=1}^{N}\frac{|S_c({\bf F}^{l}_i)-\sum_{k=1}^K\sum_{x,y}\left(w^k_c{F^{lk}_i(x,y)}\right)|}{|S_c({\bf F}^{l}_i)|},
\end{equation}
\end{small}


\rg{We report the comparison results of the different CAM methods in Table \ref{tab:tab1}. Note that, the lower value of the sensitivity and conservation indicates that the method suits the axioms better. It is clear that the Grad-CAM++ breaks the axioms of sensitivity and conservation with poor performance in the axiomatic evaluation. This may further explain why Grad-CAM++ cannot achieve comparable performance in class discrimination analysis. The results imply that it may be important to consider the axioms in designing visualization methods.}

\section{Conclusion}
In this paper, we present a novel visualization method called XGrad-CAM motivated by the axioms of sensitivity and conservation. A clear mathematical explanation is provided to \unclear{fill the gap in interpretability for CAM visualization methods.} Experimental results show that our XGrad-CAM enhances Grad-CAM in terms of sensitivity and conservation, and significantly improves the visualization performance compared with Grad-CAM. \unclear{We also give a reasonable explanation why existing methods (i.e., Grad-CAM and Ablation-CAM) can be effective from the perspective of axioms.}

\section{Acknowledgement}
This work was partially supported by the National Natural Science Foundation of China (No. 61972435) and the China Scholarship Council (CSC).

\bibliography{xgrad-new}

\clearpage
\input{supp.tex}
\end{document}

%% file: supp.tex
\begin{appendix}
\section{Proof}

\qy{In this section, we aim to demonstrate that given an arbitrary layer in ReLU-CNNs, for any class of interest, there exists a specific equation between the class score and the feature maps of the layer.}


\qy{For a ReLU-CNN which only has ReLU rectification as its nonlinearity, the following equation holds for an arbitrary layer $l$:}
\begin{equation}\label{eq13}
u^{l+1}_j=\sum_{i}\left(\frac{\partial{u_j^{l+1}}}{\partial{u_i^{l}}}{u_i^{l}}\right)+b_j^{l+1},
\end{equation}

\noindent \qy{where $u^{l}_i$ represents an unit in layer $l$, $u^{l+1}_j$ represents an unit in layer $l+1$, $\frac{\partial{u_j^{l+1}}}{\partial{u_i^{l}}}$ is the gradient of $u^{l+1}_j$ w.r.t. $u^{l}_i$, $b_j^{l+1}$ is the bias term asscociated with the unit $u^{l+1}_j$. Note that, if unit $u_j^{t}$ is an output of a ReLU or pooling layer, the corresponding bias term $b_j^{t}$ is zero.}

We then prove our statement (i.e., Eq.(5) in the main paper) using \emph{mathematical induction} \cite{kindermans2016investigating}. In the top layer $L$, the response of the $c$-th unit is exactly the class score of interest $S_c$ in the main paper, and it is easy to verify that:
\begin{equation}\label{eq14}
u_c^L=\sum_{i}\left(\frac{\partial{u_c^L}}{\partial{u_i^{L-1}}}{u_i^{L-1}}\right)+b_c^{L},
\end{equation}

Suppose that for layer $l$ ($l<L$):
\begin{equation}\label{eq15}
u_c^L=\sum_{i}\left(\frac{\partial{u_c^L}}{\partial{u_i^{l}}}{u_i^{l}}\right)+\sum_{t=l+1}^L\sum_{k}\frac{\partial{u_c^L}}{\partial{u_k^{t}}}b_k^{t},
\end{equation}

Then, for layer $l-1$, it holds:
\begin{equation}\label{eq16}
\begin{split}
&\sum_{i^{'}}\left(\frac{\partial{u_c^L}}{\partial{u_{i^{'}}^{l-1}}}{u_{i^{'}}^{l-1}}\right)+\sum_{t=l}^L\sum_{k^{'}}\frac{\partial{u_c^L}}{\partial{u_{k^{'}}^{t}}}b_{k^{'}}^{t}\\
&=\sum_{i^{'}}\left(\frac{\partial{u_c^L}}{\partial{u_{i^{'}}^{l-1}}}{u_{i^{'}}^{l-1}}\right)+\sum_{k^{'}}\frac{\partial{u_c^L}}{\partial{u_{k^{'}}^{l}}}b_{k^{'}}^{l}+\sum_{t=l+1}^L\sum_{k}\frac{\partial{u_c^L}}{\partial{u_{k}^{t}}}b_k^{t}\\
&=\sum_{i^{'}}\left(\sum_{i}\left(\frac{\partial{u_c^L}}{\partial{u_{i}^{l}}}\frac{\partial{u_{i}^{l}}}{\partial{u_{i^{'}}^{l-1}}}\right){u_{i^{'}}^{l-1}}\right)+\sum_{k^{'}}\frac{\partial{u_c^L}}{\partial{u_{k^{'}}^{l}}}b_{k^{'}}^{l}+\sum_{t=l+1}^L\sum_{k}\frac{\partial{u_c^L}}{\partial{u_{k}^{t}}}b_k^{t}\\
&=\sum_{i}\frac{\partial{u_c^L}}{\partial{u_{i}^{l}}}\left(\sum_{i^{'}}\left(\frac{\partial{u_{i}^{l}}}{\partial{u_{i^{'}}^{l-1}}}{u_{i^{'}}^{l-1}}\right)\right)+\sum_{k^{'}}\frac{\partial{u_c^L}}{\partial{u_{k^{'}}^{l}}}b_{k^{'}}^{l}+\sum_{t=l+1}^L\sum_{k}\frac{\partial{u_c^L}}{\partial{u_{k}^{t}}}b_k^{t}\\
&=\sum_{i}\frac{\partial{u_c^L}}{\partial{u_{i}^{l}}}\left(\sum_{i^{'}}\left(\frac{\partial{u_{i}^{l}}}{\partial{u_{i^{'}}^{l-1}}}{u_{i^{'}}^{l-1}}\right)+b_{i}^{l}\right)+\sum_{t=l+1}^L\sum_{k}\frac{\partial{u_c^L}}{\partial{u_{k}^{t}}}b_k^{t}\\
&=\sum_{i}\left(\frac{\partial{u_c^L}}{\partial{u_{i}^{l}}}{u_{i}^{l}}\right)+\sum_{t=l+1}^L\sum_{k}\frac{\partial{u_c^L}}{\partial{u_k^{t}}}b_k^{t}\end{split}
\end{equation}
i.e.,
\begin{equation}\label{eq17}
u_c^L=\sum_{i^{'}}\left(\frac{\partial{u_c^L}}{\partial{u_{i^{'}}^{l-1}}}{u_{i^{'}}^{l-1}}\right)+\sum_{t=l}^L\sum_{k^{'}}\frac{\partial{u_c^L}}{\partial{u_{k^{'}}^{t}}}b_{k^{'}}^{t},
\end{equation}

\qy{This means that for an arbitrary layer, the class score equals to the sum of
gradient$\times$feature plus an extra bias term.}

\section{$\epsilon({\bf F}^l)$ and $\zeta({\bf F}^{l};k)$}
\begin{figure}
\setlength{\belowcaptionskip}{-0.5cm}
\begin{tabular}{cc}\hspace{1.4mm}
\bmvaHangBox{\fbox{\parbox{5.2cm}{\includegraphics[width=5.2cm]{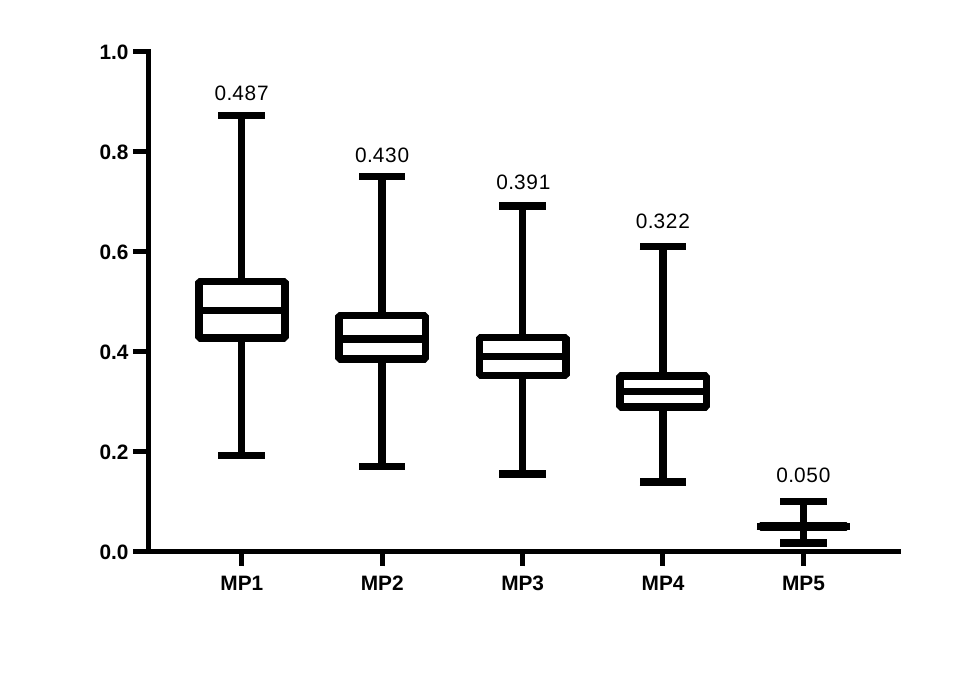}}}}&
\bmvaHangBox{\fbox{\parbox{5.8cm}{\includegraphics[width=5.8cm]{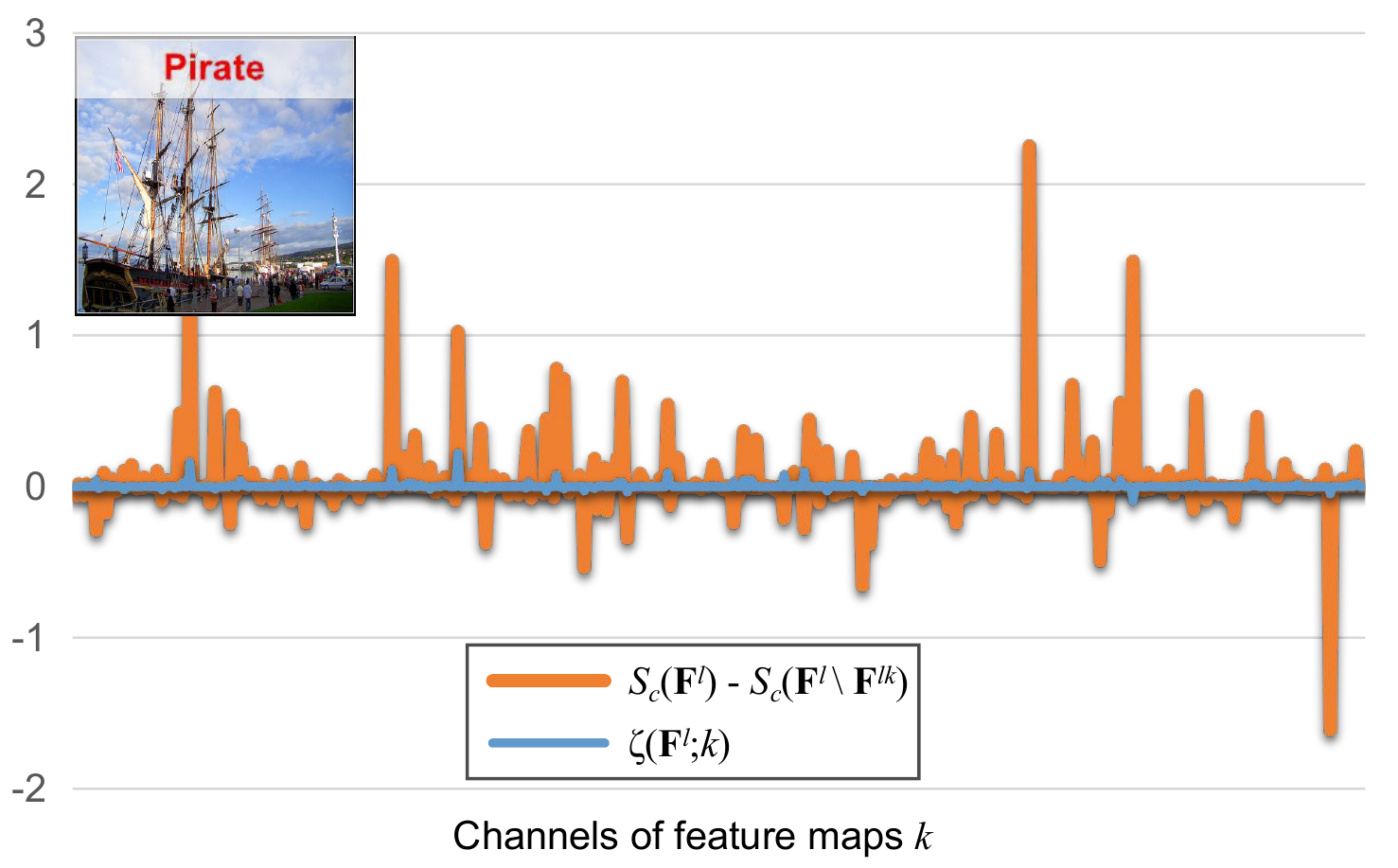}}}}\\
(a)&(b)
\end{tabular}
\caption{(a) Normalized $\epsilon({\bf F}^l)$ in different layers of VGG-16 model. ``MP'' represents for Maxpooling layer. The mean values are provided above the box-plots; (b) $\frac{\left|\zeta({\bf F}^{l};k)\right|}{\left|S_c({\bf F}^{l})-S_c({\bf F}^{l}\backslash{\bf F}^{lk})\right|}$ is small for most of the feature maps. Exceptions usually happen in the unimportant feature maps.}
\label{fig:fig8}
\end{figure}

$\epsilon({\bf F}^l)$ is the bias term in Eq. (5) in the main paper. We calculated $\left|\frac{\epsilon({\bf F}^l)}{S_c({\bf F}^l)}\right|$ of 1000 input images in different layers of VGG-16 model, with the class of interest $c$ set as the top-1 predicted class. Fig. ~\ref{fig:fig8}(a) shows that this term is rather large in shallow layers.


$\zeta({\bf F}^{l};k)$ is a bias term in Eq. (6) in the main paper. Given an input example, Fig. ~\ref{fig:fig8}(b) shows the values of $S_c({\bf F}^{l})-S_c({\bf F}^{l}\backslash{\bf F}^{lk})$ and $\zeta({\bf F}^{l};k)$ w.r.t. all the feature maps in the last spatial layer of VGG16 model. It can be seen that $\frac{\left|\zeta({\bf F}^{l};k)\right|}{\left|S_c({\bf F}^{l})-S_c({\bf F}^{l}\backslash{\bf F}^{lk})\right|}$ is rather small for most of the feature maps. Exceptions usually happen in the unimportant feature maps whose removing only lead to a tiny score change.

\section{CAM, Grad-CAM, Ablation-CAM and XGrad-CAM on GAP-CNNs}

In this section, we prove that for GAP-CNNs (e.g., ResNet-101, Inception\_v3), CAM \cite{Zhou2016Learning}, Grad-CAM \cite{Selvaraju2017Grad}, Ablation-CAM \cite{ramaswamy2020ablation} and our XGrad-CAM achieve the same performance on the last spatial layers of the models.

GAP-CNNs usually consist of fully-convolution layers, global average pooling and a linear classifier with softmax. Specifically, let ${\bf F}^{l}$ be the last spatial layer, the output of the global average pooling is:

\begin{equation}\label{eq18}
A^k=\frac{1}{Z}\sum_{x,y}F^{lk}(x,y)
\end{equation}
where $Z$ is the number of units in the $k$-th feature map. The score of class $c$ is exactly a weighted sum of $A^k$ since the classifier is linear:

\begin{equation}\label{eq19}
S_c=\sum_{k=1}^K\left(w_c^kA^k\right)+b_c
\end{equation}
where $w_c^k$ is the weight connecting the $k$-th feature map with the $c$-th class, $b_c$ is a bias. Combining Eq. \eqref{eq18} and Eq. \eqref{eq19}, we have:

\begin{equation}\label{eq20}
\begin{split}
S_c({\bf F}^{l})&=\frac{1}{Z}\sum_{x,y}\sum_{k=1}^K\left(w_c^kF^{lk}(x,y)\right) +b_c
\end{split}
\end{equation}

The weight of CAM \cite{Zhou2016Learning} is then defined as $w_c^k$.

For a GAP-CNN, we can simply get that $\forall{x,y}$, $\frac{\partial{S_c({\bf F}^{l})}}{\partial{F^{lk}(x,y)}}=\frac{1}{Z}w_c^k$ using the Chain Rule. Recall the definition of the weights in Grad-CAM \cite{Selvaraju2017Grad}, we have:
\begin{equation}\label{eq21}
\frac{1}{Z}\sum_{x,y}\frac{\partial{S_c({\bf F}^{l})}}{\partial{F^{lk}(x,y)}}=\frac{1}{Z}w_c^k.
\end{equation}

Recall the definition of the weights in Ablation-CAM \cite{ramaswamy2020ablation}, we have:
\begin{equation}\label{eq22}
\frac{S_c({\bf F}^{l})-S_c({\bf F}^{l}\backslash{\bf F}^{lk})}{\sum_{x,y}F^{lk}(x,y)}=\frac{w_c^kA^k}{ZA^k}=\frac{1}{Z}w_c^k.
\end{equation}

Recall the definition of the weights in XGrad-CAM, we have:

\begin{small}
\begin{equation}\label{eq23}
\sum_{x,y}\left(\frac{F^{lk}(x,y)}{\sum_{x,y}{F^{lk}(x,y)}}\frac{\partial{S_c({\bf F}^{l})}}{\partial{F^{lk}(x,y)}}\right)=\sum_{x,y}\left(\frac{F^{lk}(x,y)}{\sum_{x,y}{F^{lk}(x,y)}}\frac{1}{Z}w_c^k\right)=\frac{1}{Z}w_c^k.
\end{equation}
\end{small}

It shows that the weights of Grad-CAM \cite{Selvaraju2017Grad}, Ablation-CAM \cite{ramaswamy2020ablation} and XGrad-CAM are exactly the same in the case of GAP-CNNs. Besides, they are also identical to the weight of CAM \cite{Zhou2016Learning} expect a constant $Z$, which makes no difference for visualization.
Therefore, we can conclude that CAM \cite{Zhou2016Learning}, Grad-CAM \cite{Selvaraju2017Grad}, Ablation-CAM \cite{ramaswamy2020ablation} and XGrad-CAM achieve the same performance on the last spatial layers of GAP-CNNs.

\section{Additional Visualization Results}
\begin{figure*}[tb]
\begin{center}
\fbox{\parbox{11cm}{\includegraphics[width=11cm]{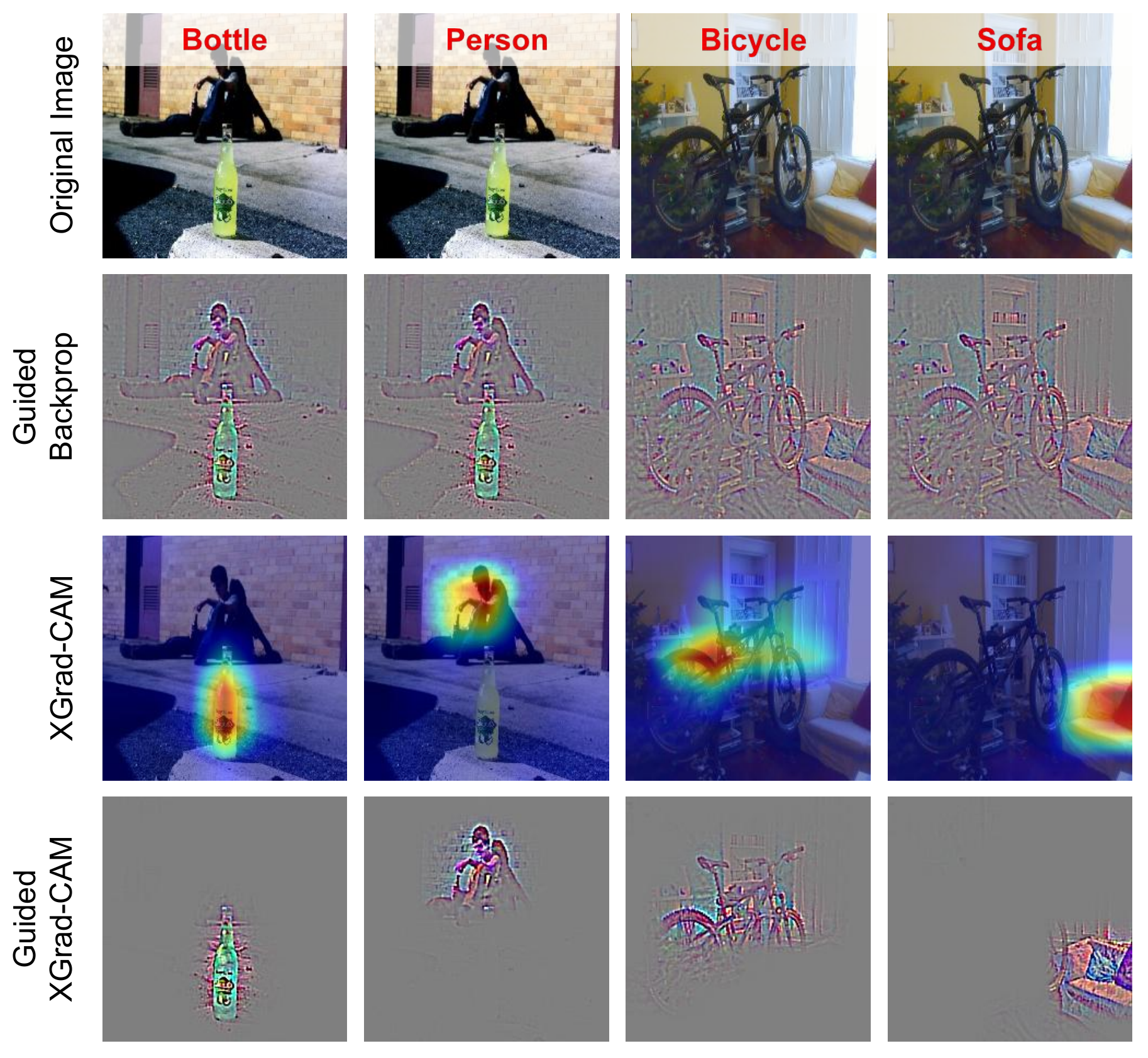}}}
\end{center}
\caption{Several visualization results of XGrad-CAM, Guided Backprop and Guided XGrad-CAM.}
\label{fig:fig11}
\end{figure*}

In the section of class discrimination analysis in the main paper, we evaluated the class-discriminability of different CAM methods using their guided versions rather than themselves. The motivation comes from two aspects. First, the guided versions have the same class-discriminability as the original versions. As shown in Fig. \ref{fig:fig11}, we visualized several visualization results of XGrad-CAM, Guided Backprop \cite{Springenberg2014Striving} and Guided XGrad-CAM. It is shown that Guided XGrad-CAM inherits the class-discriminability of XGrad-CAM completely. This phenomenon applies to all the other CAM methods. Second, the results of guided versions provides a better visualization for the objects of interest with more object details. It helps the subjects make their decisions more accurately and efficiently in the game of  ``What do you see'' as shown in Fig.4(a) in the main paper.

\begin{figure*}[tb]
\begin{center}
\fbox{\parbox{12.5cm}{\includegraphics[width=12.5cm]{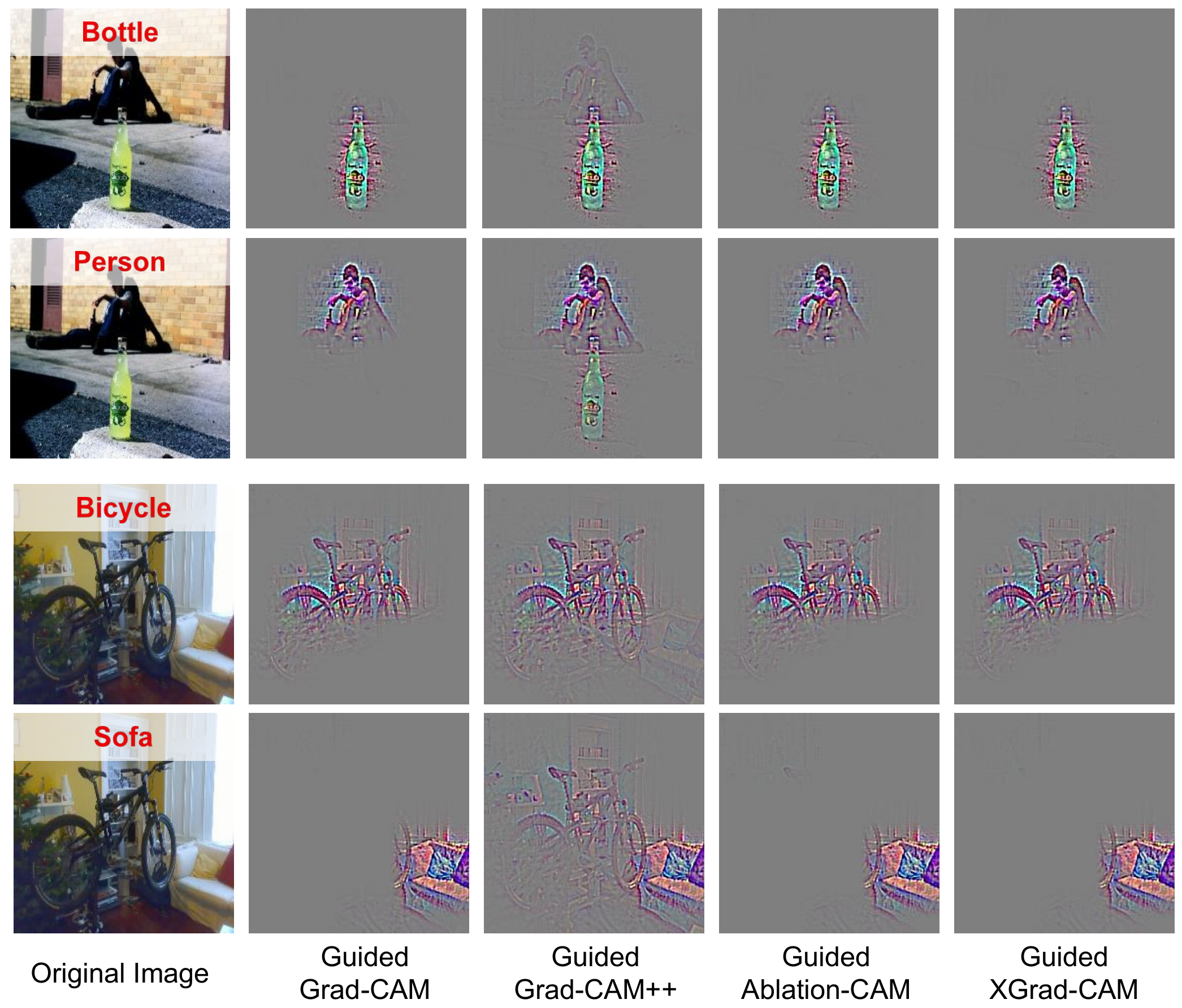}}}
\end{center}
\caption{Additional visualization results to compare the class-discriminability of different CAM methods.}
\label{fig:fig7}
\end{figure*}

Fig. \ref{fig:fig7} presents several qualitative results in VOC 2007 validation set to further compare the class-discriminability of different CAM methods. We can see that if there are objects belonging to multiple classes in an image, Grad-CAM++ also highlights regions of irrelevant classes. Clearly, Grad-CAM++ is not class-discriminative compared with the other three CAM methods.

\end{appendix}

%% file: main_new.bbl
\begin{thebibliography}{35}
\providecommand{\natexlab}[1]{#1}
\providecommand{\url}[1]{\texttt{#1}}
\expandafter\ifx\csname urlstyle\endcsname\relax
  \providecommand{\doi}[1]{doi: #1}\else
  \providecommand{\doi}{doi: \begingroup \urlstyle{rm}\Url}\fi

\bibitem[Bach et~al.(2015)Bach, Binder, Montavon, Klauschen, M{\"u}ller, and
  Samek]{bach2015pixel}
Sebastian Bach, Alexander Binder, Gr{\'e}goire Montavon, Frederick Klauschen,
  Klaus-Robert M{\"u}ller, and Wojciech Samek.
\newblock On pixel-wise explanations for non-linear classifier decisions by
  layer-wise relevance propagation.
\newblock \emph{PloS one}, 10\penalty0 (7):\penalty0 e0130140, 2015.

\bibitem[Bau et~al.(2017)Bau, Zhou, Khosla, Oliva, and
  Torralba]{bau2017network}
David Bau, Bolei Zhou, Aditya Khosla, Aude Oliva, and Antonio Torralba.
\newblock Network dissection: Quantifying interpretability of deep visual
  representations.
\newblock In \emph{CVPR}, pages 3319--3327, 2017.

\bibitem[Chattopadhay et~al.(2018)Chattopadhay, Sarkar, Howlader, and
  Balasubramanian]{Aditya2017Grad}
Aditya Chattopadhay, Anirban Sarkar, Prantik Howlader, and Vineeth~N
  Balasubramanian.
\newblock Grad-{CAM}++: Generalized gradient-based visual explanations for deep
  convolutional networks.
\newblock In \emph{WACV}, pages 839--847, 2018.

\bibitem[Desai and Ramaswamy(2020)]{ramaswamy2020ablation}
Saurabh Desai and Harish~G Ramaswamy.
\newblock Ablation-cam: Visual explanations for deep convolutional network via
  gradient-free localization.
\newblock In \emph{WACV}, pages 972--980, 2020.

\bibitem[Dong et~al.(2019)Dong, Su, Wu, Li, Liu, Zhang, and
  Zhu]{dong2019efficient}
Yinpeng Dong, Hang Su, Baoyuan Wu, Zhifeng Li, Wei Liu, Tong Zhang, and Jun
  Zhu.
\newblock Efficient decision-based black-box adversarial attacks on face
  recognition.
\newblock In \emph{CVPR}, pages 7714--7722, 2019.

\bibitem[Fong and Vedaldi(2017)]{fong2017interpretable}
Ruth~C Fong and Andrea Vedaldi.
\newblock Interpretable explanations of black boxes by meaningful perturbation.
\newblock In \emph{ICCV}, pages 3429--3437, 2017.

\bibitem[Girshick et~al.(2014)Girshick, Donahue, Darrell, and
  Malik]{Girshick2013Rich}
Ross Girshick, Jeff Donahue, Trevor Darrell, and Jitendra Malik.
\newblock Rich feature hierarchies for accurate object detection and semantic
  segmentation.
\newblock In \emph{CVPR}, pages 580--587, 2014.

\bibitem[Guidotti et~al.(2018)Guidotti, Monreale, Ruggieri, Turini, Giannotti,
  and Pedreschi]{guidotti2019survey}
Riccardo Guidotti, Anna Monreale, Salvatore Ruggieri, Franco Turini, Fosca
  Giannotti, and Dino Pedreschi.
\newblock A survey of methods for explaining black box models.
\newblock \emph{ACM computing surveys}, 51\penalty0 (5):\penalty0 1--42, 2018.

\bibitem[He et~al.(2016)He, Zhang, Ren, and Sun]{He2016Identity}
Kaiming He, Xiangyu Zhang, Shaoqing Ren, and Jian Sun.
\newblock Identity mappings in deep residual networks.
\newblock In \emph{ECCV}, pages 630--645, 2016.

\bibitem[He et~al.(2017)He, Gkioxari, Doll{\'a}r, and
  Girshick]{Kaiming2017Mask}
Kaiming He, Georgia Gkioxari, Piotr Doll{\'a}r, and Ross Girshick.
\newblock Mask r-cnn.
\newblock In \emph{ICCV}, pages 2961--2969, 2017.

\bibitem[Kindermans et~al.(2016)Kindermans, Sch{\"u}tt, M{\"u}ller, and
  D{\"a}hne]{kindermans2016investigating}
Pieter-Jan Kindermans, Kristof Sch{\"u}tt, Klaus-Robert M{\"u}ller, and Sven
  D{\"a}hne.
\newblock Investigating the influence of noise and distractors on the
  interpretation of neural networks.
\newblock \emph{arXiv preprint arXiv:1611.07270}, 2016.

\bibitem[Krizhevsky et~al.(2012)Krizhevsky, Sutskever, and
  Hinton]{Krizhevsky2012ImageNet}
Alex Krizhevsky, Ilya Sutskever, and Geoffrey~E Hinton.
\newblock Imagenet classification with deep convolutional neural networks.
\newblock In \emph{NeurIPS}, pages 1097--1105, 2012.

\bibitem[Li et~al.(2019)Li, Schmidt, and Kolter]{li2019adversarial}
Juncheng Li, Frank Schmidt, and Zico Kolter.
\newblock Adversarial camera stickers: A physical camera-based attack on deep
  learning systems.
\newblock In \emph{ICML}, pages 3896--3904, 2019.

\bibitem[Long et~al.(2015)Long, Shelhamer, and Darrell]{long2015fully}
Jonathan Long, Evan Shelhamer, and Trevor Darrell.
\newblock Fully convolutional networks for semantic segmentation.
\newblock In \emph{CVPR}, pages 3431--3440, 2015.

\bibitem[Mahendran and Vedaldi(2016)]{mahendran2016visualizing}
Aravindh Mahendran and Andrea Vedaldi.
\newblock Visualizing deep convolutional neural networks using natural
  pre-images.
\newblock \emph{International Journal of Computer Vision}, 120\penalty0
  (3):\penalty0 233--255, 2016.

\bibitem[Montavon et~al.(2017)Montavon, Lapuschkin, Binder, Samek, and
  M{\"u}ller]{montavon2017explaining}
Gr{\'e}goire Montavon, Sebastian Lapuschkin, Alexander Binder, Wojciech Samek,
  and Klaus-Robert M{\"u}ller.
\newblock Explaining nonlinear classification decisions with deep taylor
  decomposition.
\newblock \emph{Pattern Recognition}, 65:\penalty0 211--222, 2017.

\bibitem[Montavon et~al.(2018)Montavon, Samek, and
  M{\"u}ller]{montavon2018methods}
Gr{\'e}goire Montavon, Wojciech Samek, and Klaus-Robert M{\"u}ller.
\newblock Methods for interpreting and understanding deep neural networks.
\newblock \emph{Digital Signal Processing}, 73:\penalty0 1--15, 2018.

\bibitem[Omeiza et~al.(2019)Omeiza, Speakman, Cintas, and
  Weldermariam]{omeiza2019smooth}
Daniel Omeiza, Skyler Speakman, Celia Cintas, and Komminist Weldermariam.
\newblock Smooth grad-cam++: An enhanced inference level visualization
  technique for deep convolutional neural network models.
\newblock \emph{arXiv preprint arXiv:1908.01224}, 2019.

\bibitem[Paszke et~al.(2019)Paszke, Gross, Massa, Lerer, Bradbury, Chanan,
  Killeen, Lin, Gimelshein, Antiga, et~al.]{paszke2017automatic}
Adam Paszke, Sam Gross, Francisco Massa, Adam Lerer, James Bradbury, Gregory
  Chanan, Trevor Killeen, Zeming Lin, Natalia Gimelshein, Luca Antiga, et~al.
\newblock Pytorch: An imperative style, high-performance deep learning library.
\newblock In \emph{NeurIPS}, pages 8026--8037, 2019.

\bibitem[Qin et~al.(2018)Qin, Yu, Liu, and Chen]{qin2018convolutional}
Zhuwei Qin, Fuxun Yu, Chenchen Liu, and Xiang Chen.
\newblock How convolutional neural networks see the world---a survey of
  convolutional neural network visualization methods.
\newblock \emph{Mathematical Foundations of Computing}, 1\penalty0
  (2):\penalty0 149--180, 2018.

\bibitem[Ren et~al.(2015)Ren, He, Girshick, and Sun]{Ren2015Faster}
Shaoqing Ren, Kaiming He, Ross Girshick, and Jian Sun.
\newblock Faster r-cnn: Towards real-time object detection with region proposal
  networks.
\newblock In \emph{NeurIPS}, pages 91--99, 2015.

\bibitem[Russakovsky et~al.(2015)Russakovsky, Deng, Su, Krause, Satheesh, Ma,
  Huang, Karpathy, Khosla, Bernstein, et~al.]{russakovsky2015imagenet}
Olga Russakovsky, Jia Deng, Hao Su, Jonathan Krause, Sanjeev Satheesh, Sean Ma,
  Zhiheng Huang, Andrej Karpathy, Aditya Khosla, Michael Bernstein, et~al.
\newblock Imagenet large scale visual recognition challenge.
\newblock \emph{International Journal of Computer Vision}, 115\penalty0
  (3):\penalty0 211--252, 2015.

\bibitem[Samek et~al.(2019)Samek, Montavon, Vedaldi, Hansen, and
  M{\"u}ller]{samek2019explainable}
Wojciech Samek, Gr{\'e}goire Montavon, Andrea Vedaldi, Lars~Kai Hansen, and
  Klaus-Robert M{\"u}ller.
\newblock \emph{Explainable AI: interpreting, explaining and visualizing deep
  learning}, volume 11700.
\newblock Springer Nature, 2019.

\bibitem[Selvaraju et~al.(2017)Selvaraju, Cogswell, Das, Vedantam, Parikh, and
  Batra]{Selvaraju2017Grad}
Ramprasaath~R Selvaraju, Michael Cogswell, Abhishek Das, Ramakrishna Vedantam,
  Devi Parikh, and Dhruv Batra.
\newblock Grad-{CAM}: Visual explanations from deep networks via gradient-based
  localization.
\newblock In \emph{ICCV}, pages 618--626, 2017.

\bibitem[Simonyan and Zisserman(2014)]{Simonyan2014Very}
Karen Simonyan and Andrew Zisserman.
\newblock Very deep convolutional networks for large-scale image recognition.
\newblock \emph{arXiv preprint arXiv:1409.1556}, 2014.

\bibitem[Simonyan et~al.(2013)Simonyan, Vedaldi, and
  Zisserman]{simonyan2013deep}
Karen Simonyan, Andrea Vedaldi, and Andrew Zisserman.
\newblock Deep inside convolutional networks: Visualising image classification
  models and saliency maps.
\newblock \emph{arXiv preprint arXiv:1312.6034}, 2013.

\bibitem[Smilkov et~al.(2017)Smilkov, Thorat, Kim, Vi{\'e}gas, and
  Wattenberg]{smilkov2017smoothgrad}
Daniel Smilkov, Nikhil Thorat, Been Kim, Fernanda Vi{\'e}gas, and Martin
  Wattenberg.
\newblock Smoothgrad: removing noise by adding noise.
\newblock \emph{arXiv preprint arXiv:1706.03825}, 2017.

\bibitem[Springenberg et~al.(2014)Springenberg, Dosovitskiy, Brox, and
  Riedmiller]{Springenberg2014Striving}
Jost~Tobias Springenberg, Alexey Dosovitskiy, Thomas Brox, and Martin
  Riedmiller.
\newblock Striving for simplicity: The all convolutional net.
\newblock \emph{arXiv preprint arXiv:1412.6806}, 2014.

\bibitem[Su et~al.(2019)Su, Vargas, and Sakurai]{su2019one}
Jiawei Su, Danilo~Vasconcellos Vargas, and Kouichi Sakurai.
\newblock One pixel attack for fooling deep neural networks.
\newblock \emph{IEEE Transactions on Evolutionary Computation}, 2019.

\bibitem[Sundararajan et~al.(2017)Sundararajan, Taly, and
  Yan]{sundararajan2017axiomatic}
Mukund Sundararajan, Ankur Taly, and Qiqi Yan.
\newblock Axiomatic attribution for deep networks.
\newblock In \emph{ICML}, pages 3319--3328, 2017.

\bibitem[Szegedy et~al.(2016)Szegedy, Vanhoucke, Ioffe, Shlens, and
  Wojna]{Szegedy2016Rethinking}
Christian Szegedy, Vincent Vanhoucke, Sergey Ioffe, Jon Shlens, and Zbigniew
  Wojna.
\newblock Rethinking the inception architecture for computer vision.
\newblock In \emph{CVPR}, pages 2818--2826, 2016.

\bibitem[Zeiler and Fergus(2014)]{zeiler2014visualizing}
Matthew~D Zeiler and Rob Fergus.
\newblock Visualizing and understanding convolutional networks.
\newblock In \emph{ECCV}, pages 818--833, 2014.

\bibitem[Zhou et~al.(2016)Zhou, Khosla, Lapedriza, Oliva, and
  Torralba]{Zhou2016Learning}
Bolei Zhou, Aditya Khosla, Agata Lapedriza, Aude Oliva, and Antonio Torralba.
\newblock Learning deep features for discriminative localization.
\newblock In \emph{CVPR}, pages 2921--2929, 2016.

\bibitem[Zhou and Kainz(2018)]{zhou2018efficient}
Keyang Zhou and Bernhard Kainz.
\newblock Efficient image evidence analysis of cnn classification results.
\newblock \emph{arXiv preprint arXiv:1801.01693}, 2018.

\bibitem[Zintgraf et~al.(2017)Zintgraf, Cohen, Adel, and
  Welling]{zintgraf2017visualizing}
Luisa~M Zintgraf, Taco~S Cohen, Tameem Adel, and Max Welling.
\newblock Visualizing deep neural network decisions: Prediction difference
  analysis.
\newblock \emph{arXiv preprint arXiv:1702.04595}, 2017.

\end{thebibliography}
